\documentclass[journal]{IEEEtran}
\usepackage{amsmath,amsfonts}
\usepackage{array}
\usepackage[caption=false,font=normalsize,labelfont=sf,textfont=sf]{subfig}
\usepackage{textcomp}
\usepackage{stfloats}
\usepackage{url}
\usepackage{verbatim}
\usepackage{graphicx}
\usepackage{cite}
\usepackage{multirow, multicol}
\usepackage{tabularx}
\usepackage{booktabs}
\usepackage{bm}
\usepackage[linesnumbered,ruled]{algorithm2e}
\usepackage{orcidlink}
\hypersetup{
            colorlinks=true,
            linkcolor=blue,
            anchorcolor=blue,
            citecolor=blue
            }

\begin{document}

% \title{A Sample Article Using IEEEtran.cls\\ for IEEE Journals and Transactions}

\title{FullTransNet: Full Transformer with Local-Global Attention for Video Summarization}
\author{Libin Lan$^{\orcidlink{0000-0003-4754-813X}}$, \IEEEmembership{Member, IEEE}, Lu Jiang$^{\orcidlink{0009-0007-2686-9133}}$, Tianshu Yu$^{\orcidlink{0000-0002-6537-1924}}$, Xiaojuan Liu$^{\orcidlink{0000-0002-9178-2006}}$, and Zhongshi He$^{\orcidlink{0000-0002-6907-5743}}$
        <-this % stops a space
\thanks{This paragraph of the first footnote will contain the date on which you submitted your paper for review. This work was supported in part by the Scientific Research Foundation of Chongqing University of Technology under Grants 0103210650 and 0121230235 and in part by the Youth Project of Science and Technology Research Program of Chongqing Education Commission of China under Grants KJQN202301145 and KJQN202301162. \textit{(Corresponding authors: Libin Lan.)}}
\thanks{Libin Lan and Lu Jiang are with the College of Computer Science and Engineering, Chongqing University of Technology, Chongqing 400054, China (e-mail: lanlbn@cqut.edu.cn; bdml\_jl@stu.cqut.edu.cn).}
\thanks{Tianshu Yu is with the School of Data Science, Chinese University of Hong Kong, Shenzhen 518172, China (e-mail: yutianshu@cuhk.edu.cn).}
\thanks{Xiaojuan Liu is with the College of Artificial Intelligence, Chongqing University of Technology, Chongqing 401135, China (e-mail: liuxiaojuan0127@cqut.edu.cn).}
\thanks{Zhongshi He is with the College of Computer Science, Chongqing University, Chongqing 400044, China (e-mail: zshe@cqu.edu.cn).}
}

\markboth{Journal of \LaTeX\ Class Files,~Vol.~14, No.~8, August~2021}%
% \markboth{IEEE Transactions on Neural Networks and Learning Systems,~Vol.~14, No.~8, August~2021}%
{Lan \MakeLowercase{\textit{et al.}}: FullTransNet: Full Transformer with Local-Global Attention for Video Summarization}


\maketitle

% \begin{abstract}
% This document describes the most common article elements and how to use the IEEEtran class with \LaTeX \ to produce files that are suitable for submission to the IEEE.  IEEEtran can produce conference, journal, and technical note (correspondence) papers with a suitable choice of class options. 
% \end{abstract}

\begin{abstract}
Video summarization aims to generate a compact, informative, and representative synopsis of raw videos, which is crucial for browsing, analyzing, and understanding video content. Dominant approaches in video summarization primarily rely on recurrent or convolutional neural networks, and more recently on encoder-only transformer architectures. However, these methods typically suffer from several limitations in parallelism, modeling long-range dependencies, and providing explicit generative capabilities. To address these issues, we propose a transformer-like architecture named FullTransNet with two-fold ideas. First, it uses a full transformer with an encoder-decoder structure as an alternative architecture for video summarization. As the full transformer is specifically designed for sequence transduction tasks, its direct application to video summarization is both intuitive and effective. Second, it replaces the standard full attention mechanism with a combination of local and global sparse attention, enabling the model to capture long-range dependencies while significantly reducing computational costs. This local-global sparse attention is applied exclusively at the encoder side, where the majority of computations occur, further enhancing efficiency. Extensive experiments on two widely used benchmark datasets, SumMe and TVSum, demonstrate that our model achieves F-scores of 54.4\% and 63.9\%, respectively, while maintaining relatively low computational and memory requirements. These results surpass the second-best performing methods by 0.1\% and 0.3\%, respectively, verifying the effectiveness and efficiency of FullTransNet. 
Our code is available on \href{https://github.com/ChiangLu/FullTransNet}{GitHub}.
\end{abstract}

% \begin{IEEEkeywords}
% Article submission, IEEE, IEEEtran, journal, \LaTeX, paper, template, typesetting.
% \end{IEEEkeywords}

\begin{IEEEkeywords}
FullTransNet, sparse attention, transformer, video summarization.
\end{IEEEkeywords}

\section{Introduction}
\label{sec:introduction}
\IEEEPARstart{V}{ideo} summarization aims to condense a video by selecting the most informative parts to create a concise summary that effectively represents the original content. The resulting summary can be presented either as a static storyboard, consisting of a set of keyframes, or as a dynamic video skim, composed of key shots \cite{Zhang2016video}. In this work, we focus on key-shot-based video skims for three primary reasons. First, viewers are more interested in watching video skims than static storyboards. Second, key shots contain diverse information that effectively represents the original video. Third, in practice, videos are usually segmented into continuous, non-overlapping shots, which preserve intrinsic visual-temporal coherence. These characteristics ensure that the selected segments accurately reflect the overall theme and storyline of the video, providing a comfortable and entertaining user experience, even when representing only a small fraction of the entire video \cite{apostolidis2020unsupervised}, \cite{zhao2017hierarchical}, \cite{zhao2018hsarnn}, \cite{li2017ageneral}, \cite{zhao2022reconstructive}, \cite{zhao2020property}. 

Considering the inherent challenges and characteristics of general video summarization tasks, it is highly significant to study single-video summarization techniques that benefit the whole video summarization community \cite{gupta2023acompre}. 
Some methods, such as VJMHT \cite{li2022joint}, TopicSum \cite{zhu2023topic}, HMT \cite{zhao2022hierarchical}, VSMHA \cite{Pang2025VSMHA}, and DGL \cite{yang2024dgl} , introduce auxiliary information from multiple videos or multimodal data to enhance summarization performance. By contrast, our approach emphasizes the ability of the summarization model to operate independently within the context of a single video, without relying on additional correlated signals or domain knowledge from multiple or domain-specific videos.
Currently, an increasing number of single-video summarization approaches have adopted advanced deep \iffalse neural \fi network architectures, such as convolutional neural networks (CNNs) \cite{rochan2018video}, \cite{li2021exploring}, recurrent neural networks (RNNs) \cite{Zhang2016video}, \cite{zhao2017hierarchical}, \cite{zhao2018hsarnn}, \cite{zhao2022reconstructive}, \cite{mahasseni2017unsupervised}, \cite{zhao2020tthrnn}, and attention mechanism-based methods \cite{apostolidis2020unsupervised}, \cite{li2021exploring}, \cite{ji2019video}, \cite{fajtl2019summarizing}, \cite{casas2018video}, \cite{liu2019learning} for video summarization. Empirical results demonstrate that these methods generally outperform traditional non-learning techniques in most cases \cite{Zhang2016video}, \cite{zhao2022reconstructive}, \cite{hsu2023video}. This is because CNNs excel at extracting spatial information from video frames, while RNNs are highly effective at modeling temporal correlations between frames. 

Although CNNs and RNNs have traditionally dominated the video summarization task, both architectures exhibit significant limitations. CNNs struggle to model long-range dependencies, primarily due to the constraints of their local receptive fields. While stacking multiple convolutional layers can help mitigate this issue, such an approach often leads to increased computational complexity. Meanwhile, RNNs inherently lack parallelism because they rely on recursive structures, where the current input depends on the previous output, making them computationally less efficient for modeling long-range dependencies. 
In contrast, the transformer architecture \cite{vaswani2017attention} leverages a self-attention mechanism to effectively capture long-range dependencies among tokens while supporting highly parallelized computation. Despite its advantages, the encoder-only transformer architecture for video summarization \cite{li2022joint}, \cite{hsu2023video} excels at understanding input sequences but lacks the explicit generative capabilities offered by a decoder.
Inspired by the success of sequence transduction tasks in natural language processing \cite{vaswani2017attention}, \cite{qu2022novelmulti}, we apply the full transformer with an encoder-decoder structure to video summarization, casting it as a sequence-to-sequence (seq2seq) learning problem. This approach is both conceptually intuitive and practically advantageous, enabling effective modeling of both the content understanding and summary generation in video summarization.

\begin{figure*}[htbp]
\centering
\includegraphics[width=1.0\linewidth, keepaspectratio]{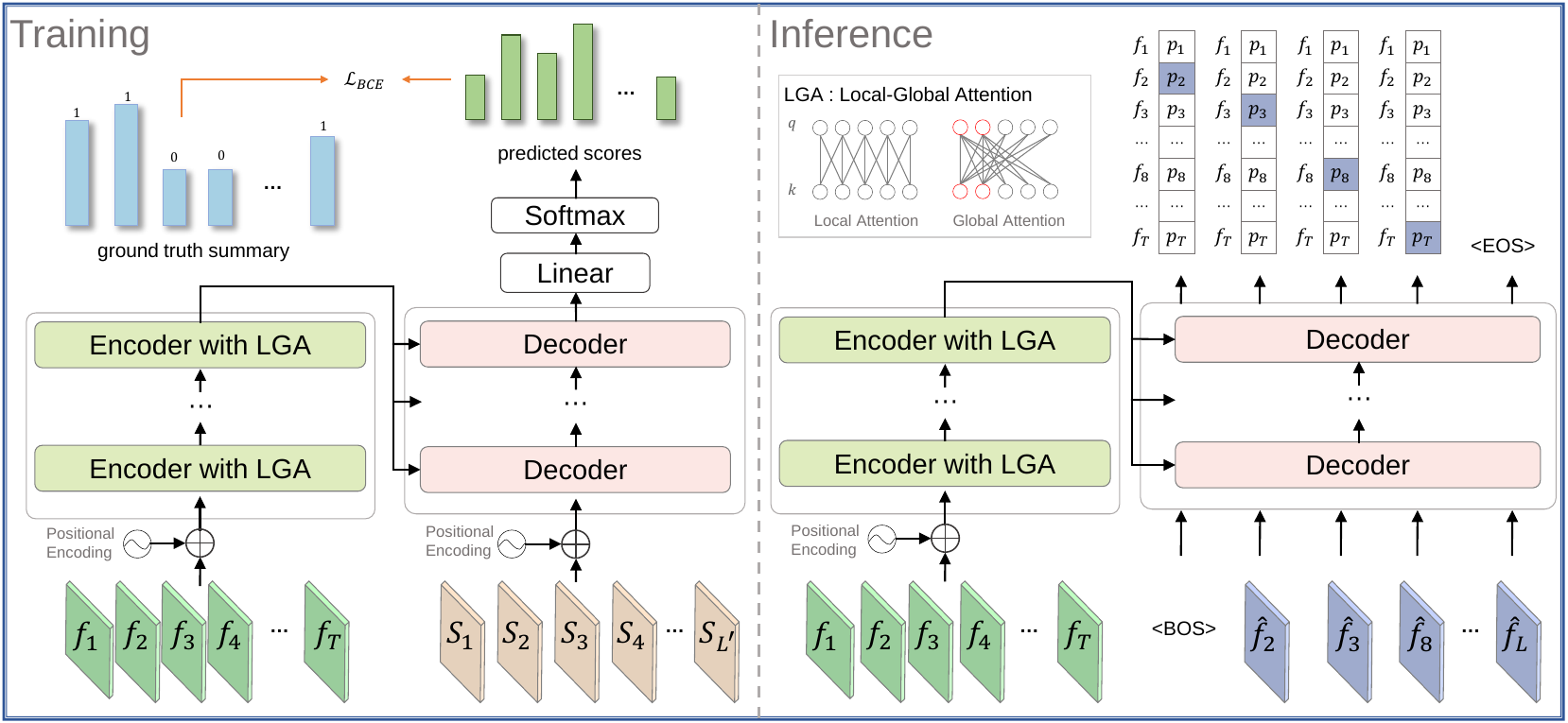}
\caption{Motivation and Main Idea. Our core idea is to directly apply full transformer architecture to video summarization and to substitute local-global attention for full attention. Our main aim is to perform video summarization in an end-to-end manner and to save computational cost without sacrificing performance. This illustration demonstrates how to implement our idea and achieve our purpose from the two processes of training and inference. Symbols $f$, $s$, $\hat{f}$, $T$, $L^\prime$ and $L$ represent frame in original video, keyshot in summary, predicted keyframe, the number of  frames in original video, the number of shots in summary, and the number of frames in summary, respectively.}
\label{fig0:motivation}
\end{figure*}

While the full transformer is effective for seq2seq modeling and excels at capturing long-range dependencies, it typically suffers from quadratic computational complexity and a heavy memory footprint relative to the sequence length. This limitation hinders its scalability and efficiency, particularly in vision tasks involving long video sequences. Therefore, it is essential to explore more efficient alternatives that maintain performance while reducing computational overhead.
To this end, our work consider replacing vanilla attention \iffalse in transformer \fi \cite{vaswani2017attention} with a sparse attention, termed \textbf{L}ocal-\textbf{G}lobal \textbf{A}ttention (LGA), which is a combination of local attention and global attention. Specifically, local attention is used to model relationships between a query and its neighbors, \iffalse neighboring tokens\fi  while global attention enables the model to attend to  important information from  the entire sequence. 
Our empirical results on video summarization tasks are consistent with prior findings in language understanding \cite{child2019generatingls}, \cite{zaheer2020bigbird}, \cite{beltagy2020longformer}, demonstrating that the sparse attention can effectively balance efficiency and performance.

In our work, the LGA is applied only on the encoder side. This decision is motivated by the fact that the decoder operates in an auto-regressive manner, where each position only needs to attend to its previous and current positions. Additionally, the length of the summary sequence is typically shorter than that of the input sequence. Consequently, using full self-attention in the decoder proves more efficient, while applying LGA in the encoder effectively reduces computational costs as sequence lengths increase.
It is worth noting that our focus is on the behavior of the full transformer and sparse attention in video summarization, but not on pushing the state-of-the-art results. Exploring more efficient attention, such as RKformer \cite{zhang2022rkformer}, ESSAformer \cite{Zhang_2023_ICCV}, and IRSAM \cite{zhang2024irsam}, could potentially lead to further improvements in performance.

Considering the aforementioned advantages of both the full transformer and sparse attention LGA, we propose a transformer-like architecture with an encoder-decoder structure, named \textbf{FullTransNet}, as an alternative solution for modeling the seq2seq problem in video summarization. Specifically, the encoder takes source sequences as input, which are the frame sequences of the original video, while the decoder processes target sequences, which correspond to ground-truth shot-level summary sequences during training or predicted sequences during inference.
\textbf{During training}, the encoder encodes the input frame sequences in parallel using self-attention, while the decoder uses cross-attention to allow each query summary frame to attend to all frames from the encoder. Additionally, masked self-attention is used in the decoder to ensure that the query frame at the current position only attends to all key-value frames up to and including this position.
\textbf{During inference}, the decoder generates summaries based on the trained FullTransNet.
\textbf{Our motivation} and \textbf{main idea} are illustrated in Fig. \ref{fig0:motivation}. Due to the powerful representation ability of full transformer, there is no need to design specific criteria to ensure that the generated summaries exhibit desirable properties such as compactness, diversity, representativeness, and informativeness \cite{Zhang2016video}, \cite{li2017ageneral}, \cite{gong2014diverse}. We train the model in a supervised learning framework, enabling the learned model to automatically produce summaries that align with human annotators' preferences.

Our main contributions are three-fold as follows:
\begin{enumerate}
    \item Our work provides new insights into generating more accurate video summaries by leveraging the full transformer architecture, marking the first attempt to apply this architecture to video summarization. Compared to encoder-only architectures, FullTransNet explicitly models both video understanding and summary generation, offering an intuitive and effective approach to video summarization while achieving promising results.
    \item Since the encoder consumes most computation, we introduce sparse attention instead of full attention into all layers of the encoder. The sparse attention combines local and global attention, aiming primarily to reduce computational burden without sacrificing performance.
    \item We conduct extensive experiments on two popular datasets, SumMe \cite{gygli2014creating} and TVSum \cite{song2015tvsum}, demonstrating the potential of using the full transformer structure and sparse attention for video summarization. Experimental results confirm that our approach achieves promising results while maintaining relatively low computational and memory demands. Notably, visualizations of the attention maps in the encoder, decoder, and encoder-decoder layers provide clear insights into how our FullTransNet works.
\end{enumerate}

The rest of this paper is organized as follows. Section \hyperref[sec:relatedwork]{\uppercase\expandafter{\romannumeral2}} delves into the findings and empirical evidence derived from prior related works. Section \hyperref[sec:method]{\uppercase\expandafter{\romannumeral3}} specifies our method, including the model’s architecture and sparse attention mechanism. Section \hyperref[sec:exp-results]{\uppercase\expandafter{\romannumeral4}} showcases the experimental results and details, particularly, the multi-layer and multi-head visualization of sparse attention mechanism. 
Section \hyperref[sec:conclusion]{\uppercase\expandafter{\romannumeral5}} summarizes our main findings and contributions, discusses the limitations of our work, and suggests possible future directions.

\section{Related work}
\label{sec:relatedwork}
Given that our approach is closely related to deep models and attention mechanisms, we review and analyze methods that apply deep learning techniques, particularly transformer architectures and their attention mechanisms, to video summarization tasks.

\subsection{Deep Model Based Video Summarization}
\subsubsection{Transformer-free Deep Model}
Transformer-free deep models primarily refer to those that rely on CNNs, RNNs, or their combinations as core building techniques for designing their respective models. The pioneering work \cite{Zhang2016video} first explores a bidirectional single-layer long short-term memory (LSTM) network for summarizing videos, achieving desirable results. Zhao et al. \cite{zhao2017hierarchical} propose a hierarchical structure to address the limitation that single-layer LSTMs struggle with long videos. The promising results of this hierarchical approach encourage Zhao et al. \cite{zhao2018hsarnn} to further refine the model by leveraging the hierarchical structure of video data. To address challenges such as large feature-to-hidden mapping matrices and long-range temporal dependencies, Zhao et al. \cite{zhao2020tthrnn} propose replacing the hierarchical structure with a tensor-train embedding layer for video summarization. Despite their success, all these methods rely on RNNs or their variants for video summarization and do not explore the application of transformers to this task.

On the other hand, some methods focus on leveraging CNNs for video summarization. Rochan et al. \cite{rochan2018video} pioneer the use of convolutional networks in video summarization by introducing a fully convolutional sequence model. This model increases the effective context size through stacked convolutional operations, enabling the network to model long-range dependencies while allowing for limited parallelization. Additionally, several works combine CNNs and RNNs for video summarization. The key idea behind this line of research is to explore how to effectively model the spatiotemporal structure of videos. For instance, Elfeki et al. \cite{elfeki2019video} integrate CNNs with gated recurrent units (GRUs) to generate spatiotemporal feature vectors, which are then used to estimate the importance of each frame. Similarly, Lal et al. \cite{lal2019nline} propose an encoder-decoder architecture using convolutional LSTMs to capture the spatiotemporal relationships among frames. Yuan et al. \cite{yuan2019spatiotemporal} leverage a combination of convolutions and LSTMs to model both the spatial and temporal structures of videos. Despite their success, these methods---whether based on CNNs, RNNs, or their combinations---face inherent limitations in parallel computing. 
Specifically, stacked CNN layers can lead to computational bottlenecks, while RNNs suffer from sequential dependencies that hinder efficient training and inference. In contrast, our work explores the application of the full Transformer architecture \cite{vaswani2017attention} to video summarization. A well-known advantage of Transformers is their high parallelism, which avoids the computational bottlenecks associated with CNNs and the sequential dependencies inherent in RNNs.

\subsubsection{Transformer-based Deep Model}
The transformer architecture is typically utilized in three ways: encoder-only, decoder-only, and full encoder-decoder, which are commonly used for understanding, generation, and transduction tasks, respectively \cite{tay2022efficient}. Currently, most published approaches applying transformers to video summarization primarily adopt the encoder-only architecture. For example, Hsu et al. \cite{hsu2023video} propose a spatiotemporal vision transformer for video summarization, which takes into account both inter-frame correlation and intra-frame attention for improved video understanding. To leverage multi-video and multimodal information (e.g., visual, textual, and audio), some works have explored multimodal transformer architectures for video summarization. Li et al. \cite{li2022joint} introduce a hierarchical transformer to explicitly model cross-video high-level semantic information for co-summarization. Similarly, Zhu et al. \cite{zhu2023topic} propose a multimodal transformer model for topic-aware video summarization, where a multimodal transformer encoder extracts features from different modalities, and a separate temporal modeling encoder captures motion dynamics. Zhao et al. \cite{zhao2022hierarchical} also propose a hierarchical multimodal transformer for video summarization, leveraging the natural structure of videos (i.e., frame-shot-video). Their model captures dependencies across frames and shots and generates summaries by exploiting scene information derived from shots.

In general, encoder-only architectures independently learn a generic representation from input data and then map this representation to the output. While this approach is advantageous for understanding input sequences, it lacks explicit generative capabilities. In contrast, full encoder-decoder architectures enable direct learning of mappings from input to output, explicitly modeling both video understanding (via the encoder) and summary generation (via the decoder). This makes the full encoder-decoder architecture a natural fit for video summarization. Additionally, inspired by the success of transformers in transduction tasks such as machine translation \cite{vaswani2017attention} and text summarization \cite{khandelwal2019SampleET}, \cite{liu2019text}, we propose employing the full transformer structure to directly learn the mapping from raw video sequences to summary sequences in an end-to-end manner.

\subsection{Attention Based video Summarization}
\subsubsection{Non-Transformer Attention}
Non-transformer attention mechanisms typically refer to those in conjunction with deep models, particularly RNNs. This type of attention focuses on relevant previous positions using an accumulated vector in the corresponding task. It has demonstrated significant advantages in natural language processing tasks, such as machine translation \cite{bahdanau2014neural}, \cite{luong2015effective}, and text summarization \cite{see2017get}, \cite{gu2016incorpation}. Additionally, it has shown promising progress in computer vision tasks, including video summarization \cite{apostolidis2020unsupervised}, \cite{ji2019video}, \cite{casas2018video}, \cite{ji2020deep}.

For transduction tasks, considering factors such as model performance, specific architecture, and application scenarios, some video summarization methods adopt customized attention mechanisms. For instance, Ji et al. \cite{ji2020deep} introduce a semantic-preserving loss with tailored attention to evaluate the decoder's output. Similarly, Casas et al. \cite{casas2018video} propose an attention mechanism to model user interest. As for the network with an encoder-decoder structure, this line of attention mechanisms is generally applied to the decoder, focusing on the decoder's state. For example, Ji et al. \cite{ji2019video}, \cite{ji2020deep} explore attention-based LSTM as a decoder to generate a sequence of importance scores. Apostolidis et al. \cite{apostolidis2020unsupervised} leverage a context attention vector from the encoder, combining it with the output of the previous time step in the decoder to reconstruct the video. Such attention mechanisms are generally computed in a sequential manner. In contrast, transformer-based attention \cite{vaswani2017attention}, i.e., self-attention, operates in a parallel fashion, offering significant computational advantages.

\subsubsection{Transformer Attention}
In this paper, we hypothesize that transformer attention, which adopts self-attention \cite{li2021exploring}, \cite{fajtl2019summarizing}, \cite{liu2019learning}, \cite{ji2020deepdistribution}, \cite{wang2020query}, \cite{Arafat2025STSA}, is characterized by its ability to perform parallel computing. Several works have applied this type of attention mechanism to video summarization. For instance, Fajtl et al. \cite{fajtl2019summarizing} employ a self-attention mechanism to perform the entire sequence-to-sequence transformation for video summarization. Ji et al. \cite{ji2020deepdistribution} incorporate a self-attention mechanism in the encoder to capture short-term contextual information. Similarly, Li et al. \cite{li2021exploring} propose a global diverse attention mechanism by adapting self-attention to estimate diverse attention weights, which are then transformed into importance scores. However, unlike these works, which involve computationally intensive components such as LSTMs, CNNs, or full attention mechanisms, our work entirely avoids LSTM and CNN operations and uses local-global sparse attention instead of full attention to train the model efficiently.

More recently, several studies have adopted standard self-attention mechanisms in video summarization tasks. For instance, Li et al. \cite{li2022joint} utilize self-attention to explicitly model high-level semantic patterns across multiple videos, aiming to improve co-summarization performance. Hsu et al. \cite{hsu2023video} incorporate both temporal and spatial attention to achieve strong summarization results. While full self-attention is effective at capturing global dependencies, it lacks an explicit inductive bias toward local structures, which can be crucial when modeling sequential or spatially correlated data such as video frames. To address this limitation and improve generalization, some works, such as C2F \cite{JIN2024104962}, attempt to incorporate sparse attention into video summarization. Similarly, we adopt a local-global sparse attention mechanism as a structural prior, ensuring that the model can be effectively trained on relatively small-scale video datasets. 
For more details on attention mechanisms in computer vision, readers can refer to related review literature \cite{tay2022efficient}, \cite{apostolidis2021Video}, \cite{guo2021attention}, \cite{LIN2022111}.

\section{Method}
\label{sec:method}
\begin{figure*}[htbp]
\centering
\includegraphics[width=1.0\linewidth, keepaspectratio]{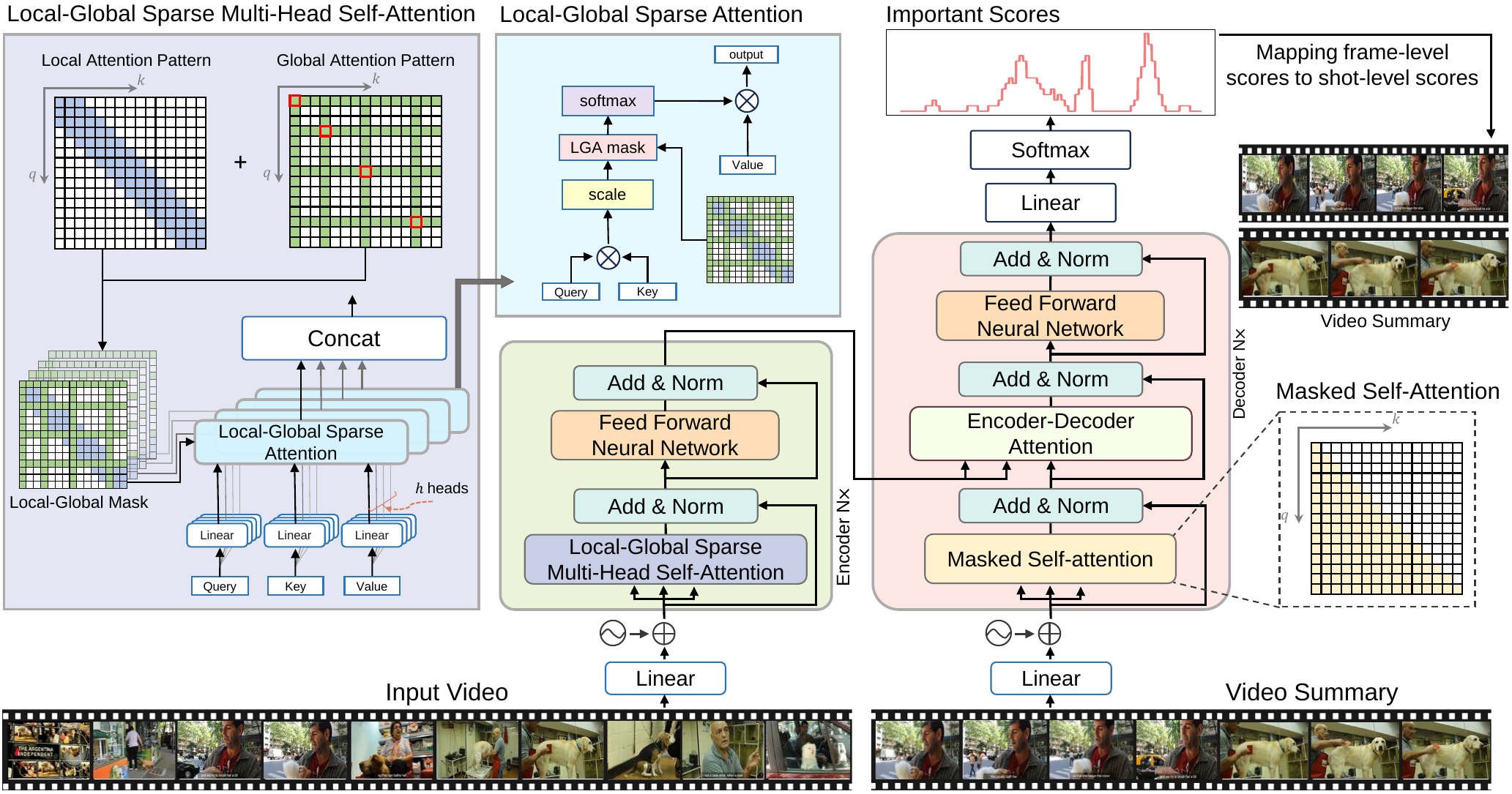}
\caption{The whole overview of FullTransNet architecture. It follows vanilla transformer with full encoder-decoder structure and exploits local-global sparse attention instead of standard full attention to build the network. The local-global sparse attention is a combination of local attention and global attention, and only used at the encoder side. Based on this sparse attention mechanism, local-global sparse multi-head attention is designed to yield a better output representation.}
\label{fig2:architecture}
\end{figure*}

Similar to \cite{ji2019video}, \cite{fajtl2019summarizing}, \cite{ji2020deepdistribution}, we formulate video summarization as a seq2seq learning problem. We address this problem using a standard transformer with a full encoder-decoder structure. In addition, taking into account the model efficiency, we use local-global sparse attention instead of full attention to balance the performance and computational cost. Our proposed model architecture, named FullTransNet, takes these two techniques as its main building blocks, as shown in Fig. \ref{fig2:architecture}. In this section, we will elaborate on each component of our model, including the encoder-decoder structure, the sparse attention mechanism, and other aspects of our approach, such as the training process. 

\subsection{Model Architecture}
\subsubsection{Encoder with Local-Global Sparse Attention}
In neural machine translation tasks, it is natural to encode input word sequences, transforming the vocabulary and structure of the input sentence into a continuous semantic representation. We apply this idea to the video summarization task, in which each video frame is considered as a word in a sentence. Thus, similar to the word sequence, the frame sequence in a video is encoded to a continuous latent representation.

We use the vanilla transformer \cite{vaswani2017attention} as our base architecture, in which the encoder is a crucial component. The encoder consists of $N$ stacked layers. Each encoder layer, except for the first one, takes as its input the output of the previous encoder layer. Each encoder layer performs a series of transformations on input sequences, progressively extracting higher-level semantic information suitable for summarization. When encoding a frame sequence, the transformer not only maps the sequence to a latent space but also can capture the dependencies between frames through the attention mechanism.

In addition, we apply linear projection to obtain embedded features from the features of each video frame extracted by GoogLeNet \cite{szegedy2015going} to keep in step with the embedding operation in the translation task and to reduce the dimensionality of video features from 1024 to $d$. Then, we incorporate positional encoding the same as one in the vanilla transformer to enable the network to perceive the spatial position information.

In brief, the encoder can be formulated as:
\begin{equation}
\label{Eq_1}
encoder(\bm{Z}_{enc}) = 
\begin{cases}
f_{enc}^{i}(\bm{Z}_{enc}+\bm{P}_{enc})\text{,} &  i=1  \\
f_{enc}^{i}(\bm{Y}_{enc}^{i-1})\text{,} &  2 \leq i \leq N 
\end{cases} \text{,}
\end{equation}
where $\bm{P}_{enc}\in\mathbb{R}^{T\times d}$ represents the positional encoding, which has the same shape as the embedded representation $\bm{Z}_{enc}$, and $f_{enc}^i$ denotes the $i$-$th$ encoder layer. $\bm{Y}_{enc}^{i-1}$ represents the output of the $(i-1)$-$th$ encoder layer. $T$ and $d$ are the number of frames in the original video and the representation dimensionality, respectively. The structure of the first encoder layer is shown in Fig. \ref{fig2:architecture} (bottom). 
Concretely, each encoder layer has two sub-layers: a multi-head self-attention (MHSA) with local-global sparse attention and a position-wise fully connected feed-forward network (FFN). In this paper, the multi-head self-attention mechanism with local-global sparse attention is referred to as \textbf{L}ocal-\textbf{G}lobal \textbf{S}parse \textbf{M}ulti-\textbf{H}ead \textbf{S}elf-\textbf{A}ttention mechanism (LGS-MHSA), which is an updated version of vanilla MHSA \cite{vaswani2017attention}.
In addition, \iffalse following the vanilla transformer \cite{vaswani2017attention},\fi a residual connection \cite{kaiming2016deep} is used around each sub-layer, followed by layer normalization \cite{ba2016layer}. Thus, the output of the two sub-layers (denoted as ${\bm{X}_i^\prime}\in\mathbb{R}^{T\times d}$ and $ {\bm{X}_i^{\prime\prime}}\in\mathbb{R}^{T\times d}$) in the $i$-th encoder can be expressed as follows:
\begin{align}
\label{Eq_23}
{\bm{X}_i^{\prime}}  &= \operatorname{LayerNorm}(\operatorname{LGS-MHSA}(\bm{X}_i\text{, }\bm{X}_i\text{, }\bm{X}_i)+\bm{X}_i) \text{,}\\
{\bm{X}_i^{\prime\prime}} &= \operatorname{LayerNorm}(\operatorname{FFN}(\bm{X}_i^{\prime})+ \bm{X}_i^{\prime}) \text{,}
\end{align}
where ${\bm{X}_i}\in\mathbb{R}^{T\times d}$ indicates the output of the $(i-1)$-$th$ encoder, i.e., $\bm{Y}_{enc}^{i-1}$.

LGS-MHSA$(\cdot)$ is an MHSA operation enhanced with local-global sparse attention. In this operation, local-global sparse attention is used to add a locality prior to the original input sequence and to designate certain positions as global tokens, enabling them to attend to all tokens in the sequence. Concerning the reasons for doing this, we will give a detailed specification and explanation in Subsection \ref{subsec:SA}.

Except for the LGS-MHSA sub-layer, an identical position-wise fully connected feed-forward network (FFN) is applied to each position. The FFN has two linear transformations with a ReLU activation in between, which is the same as the vanilla transformer \cite{vaswani2017attention}.
\begin{equation} 
\label{Eq_4}
\operatorname{FFN}(\bm{X}_i^{\prime}) = \max{(0\text{, }\bm{X}_i^{\prime }\bm{W}_1+b_1)}\bm{W}_2+b_2\text{,}
\end{equation}
where $ \bm{W}_1 $ and $ \bm{W}_2 $ represent learnable weight matrices, which are used to perform linear projections on the input data. $ b_1 $ and $ b_2 $ are bias terms, used to adjust the output after the linear projections.
\subsubsection{Decoder}
The decoder is also built upon the self-attention mechanism, but it does not employ the proposed local-global sparse attention. The whole decoder block consists of $N$ decoder layers, of which the number is identical to that of encoder layers. The input to the first decoder layer is the summary sequence along with positional encoding. Except for the first decoder layer, each layer's input includes two parts: the output from the previous decoder layer and the contextual representations from the encoder. The contextual representations contain inter-frame relationship information from the original input sequence, which is exploited by the decoder to effectively generate the target sequence. The decoder can be expressed as:
\begin{align}
\label{Eq_5}
decoder(\bm{Z}_{dec}\text{, }\bm{Y}_{enc}^{N}) = 
\begin{cases}
f_{dec}^{i}(\bm{Z}_{dec}+\bm{P}_{dec}\text{, }\bm{Y}_{enc}^{N})\text{, }  i=1 \\
f_{dec}^{i}(\bm{Y}_{dec}^{i-1}\text{, }\bm{Y}_{enc}^{N})\text{, }  2 \leq i \leq N
\end{cases}\text{,}
\end{align}
where $\bm{P}_{dec}\in\mathbb{R}^{L\times d}$ represents the positional encoding, which has the same shape as the embedded representation $\bm{Z}_{dec}$. $ \bm{Y}_{enc}^N$ means the output of the last encoder layer, and $f_{dec}^i$ denotes the $i$-$th$ decoder layer. $\bm{Y}_{dec}^{i-1}$ represents the output of the $(i-1)$-$th$ decoder layer. $L$ and $d$ are the lengths of the summary sequence formed by keyframes and the representation dimensionality, respectively. The structure of the first decoder is shown in Fig. \ref{fig2:architecture} (bottom). Concretely,  each decoder layer primarily consists of three sub-modules: masked multi-head self-attention (MMHSA), i.e., causal attention, encoder-decoder attention, i.e., cross attention (CA), and a position-wise fully connected feed-forward network (FFN). Similar to the encoder, a residual connection \cite{kaiming2016deep} is used around each sub-layer, followed by layer normalization \cite{ba2016layer}. Thus, the output of the three sub-layers (denoted as ${\bm{S}_i^\prime}\in\mathbb{R}^{L\times d}$, ${\bm{S}_i^{\prime\prime}}\in\mathbb{R}^{L\times d}$, and $ {\bm{S}_i^{\prime\prime\prime}}\in\mathbb{R}^{L\times d}$) in the $i$-th decoder can be expressed as follows:
\begin{align}
\label{Eq_678}
\bm{S}_i^{ \prime} &= \operatorname{LayerNorm}(\operatorname{MMHSA}(\bm{S}_i\text{, }\bm{S}_i\text{, }\bm{S}_i)+\bm{S}_i) \text{,}\\
\bm{S}_i^{ \prime\prime} &= \operatorname{LayerNorm}(\operatorname{CA}(\bm{Y}_{enc}^{N}\text{, }\bm{Y}_{enc}^{N}\text{, }\bm{S}_i^{\prime})+\bm{S}_i^{\prime}) \text{,}\\
\bm{S}_i^{ \prime\prime\prime}&= \operatorname{LayerNorm}(\operatorname{FFN}(\bm{S}_i^{\prime\prime})+ \bm{S}_i^{\prime\prime}) \text{,}
\end{align}
where $\bm{S}_i\in\mathbb{R}^{L\times d}$ shows the output of the $(i-1)$-$th$ decoder, i.e., $\bm{Y}_{dec}^{i-1}$. CA$(\cdot )$ receives the inputs from both the encoder and the decoder. MMHSA$( \cdot )$  allows each query $\bm{Q}$ to only attend to all the keys $\bm{K}$ and values $\bm{V}$ at the current query position and its preceding positions. This is typically implemented through a masked function applied on the unnormalized attention matrix $\hat{\bm{A}}=\operatorname{exp}(\frac{\bm{Q}\bm{K}^\top}{\sqrt d})$, which can enable parallel computation during training. All the values at the positions that do not need to be attended to are set to $-\infty$, i.e., ${\hat{\bm{A}}}_{mn}=-\infty $ if $m<n$. This masked self-attention ensures that the information after the current position is not fed to the decoder, which thus remains the auto-regressive property of the video summarization task.

Additionally, the encoder-decoder attention module allows the decoder to access the entire output of the encoder. Using the context of the input sequence can help generate the next most likely keyframe. 
The output of the last decoder layer is fed into a linear transformation layer to project its dimensionality $d$ to the length  $T$ of the original video. 
Subsequently, a softmax function layer is applied to these projected values to convert them into a probability distribution, facilitating the determination of which frames are most likely selected as keyframes.

\subsection{Local-Global Sparse Multi-Head Self-Attention}
\label{subsec:SA}

Using local-global sparse attention (LGA) instead of full attention (FA) has the two most straight purposes. First, it is used to reduce the memory footprints and computation requirements and handle longer sequences. Second, it is expected to effectively select the representative frames. 
Because the computation grows quadratically with sequence length in FA, which causes severe scalability issues, we propose using LGA to address this limitation. According to this fact that most data, particularly video data, has a locality property, directly applying softmax mapping to all tokens, i.e., assigning a weight to all tokens, seems to be paradoxical. This is because not all tokens are equally relevant or worth attending to.
Thus, it is natural to restrict each query to only attend to its neighbors. Based on this, we utilize local attention to only normalize the neighbor tokens in window $w$ from the attention matrix in a row-wise manner, so as to assign a larger weight to the tokens, which is very beneficial for selecting keyframes.

Other advantages of this LGA mechanism are also beneficial for our task. For example, LGA can introduce locality prior to mitigate the overfitting on small-scale datasets. Local operation maintains translation invariance, which makes this attention focus on similar or the same information across different positions. In addition, when multiple local window attention layers are stacked, the top layer can access all input information across the entire sequence, leading to a nearly full receptive field similar to vanilla self-attention, which makes it most likely to match vanilla self-attention in performance. Meanwhile, global attention in LGA is intentionally designed for learning the global representations across all frames in a sequence, which can alleviate the degradation of the ability to model the long-range dependencies using local attention. All these potential advantages motivate us to propose this LGA mechanism in our model architecture. The LGA is a combination of local attention and global attention. The local attention, global attention, and their combination patterns are shown in Fig. \ref{fig0:motivation} and Fig. \ref{fig2:architecture}. The detailed specifications of our LGA will be given as follows.

The attention relationships of query-key pairs, as well as the connectivity matrix between all query-key pairs, are illustrated in Fig. \ref{fig0:motivation} and Fig. \ref{fig2:architecture}. The local attention allows the model to focus on the neighbors of each frame, which helps rapidly identify keyframes. The attention computation for each query only involves elements within a small window around it, rather than the entire sequence. This means that the key-value pairs outside the window can be effectively masked out, achieving local focus and greatly reducing computation costs.

To tackle the limitations of the local attention in modeling long-range dependencies, we add the global attention to several key positions. In our configuration of LGA, the first, the middle, and the last positions are selected as key positions. The elements at key positions are treated as global tokens, which can attend to all elements in the sequence and vice versa, as illustrated in Fig. \ref{fig0:motivation} and Fig. \ref{fig2:architecture}. The global attention allows the model to learn the representations of the whole sequence, which is helpful for the decoder to predict keyframes.

Based on the above analyses, the combined sparse attention can be understood as introducing an inductive bias to the input, which is helpful to effectively train the model and save computation costs. Assuming a window size of $w$, we formulate the local-global sparse attention as the following pre-defined patterns, and then the similarity scores of query-key pairs can be computed as follows:
\begin{align}
\label{Eq_9}
\hat{\bm{A}}_{mn} = 
\begin{cases}
\frac{\bm{q}_m\bm{k}_n^\top}{\sqrt d}\text{,} &\text{if token $m$ attends to token $n$}\\
-\infty\text{,} & \text{else}
\end{cases}\text{,}
\end{align}
where $\hat{\bm{A}}_{mn}$ is unnormalized attention matrix, $m$, $n$ are indices of corresponding tokens, and \iffalse $n \in [\text{max}(0\text{, } m-\frac{w}{2})\text{, } \text{min}(n-1\text{, }m+\frac{w}{2}+1)] {\cup}\{f_{s}^{start}\text{, }f_{s}^{mid}\text{, }f_{s}^{end} \}_{s=1}^{SN}$ \fi $n \in [\text{max}(0\text{, } m-\lfloor \frac{w}{2} \rfloor)  \text{, } \text{min}(n-1\text{, }m+\lfloor \frac{w}{2} \rfloor)] {\cup}\{f_{s}^{start}\text{, }f_{s}^{mid}\text{, }f_{s}^{end} \}_{s=1}^{SN}$, $s$ is index of different shots, $SN$ is the shot number of video, $f_{s}^{start}$, $f_{s}^{mid}$, and $f_{s}^{end}$ represent the first, the middle, and the last frame of each shot. The output \iffalse representation \fi of LGA can be then computed as follows:
\begin{align}
\label{Eq_10}
\operatorname{LGA} (\hat{\bm{A}}\text{, } \bm{V})=\operatorname{softmax}(\hat{\bm{A}})\bm{V}\text{.}
\end{align}
Following \cite{vaswani2017attention}, instead of performing a single attention function LGA, Our model adopts a local-global sparse multi-head self-attention (LGS-MHSA), which allows the model to jointly attend to information from different representation subspaces at different positions. The LGS-MHSA projects $d$-dimensional queries $\bm{ Q }$, keys  $\bm{K} $, and values $\bm{V}$ to $d_k$, $d_k$, and $d_v$ dimensions by using $h$ different sets of learned projections, respectively. The LGA attention function Eq. (\ref{Eq_10}) is performed on each of the projected queries $\bm{Q}_{d_k}$, keys $\bm{K}_{d_k}$, and values $\bm{V}_{d_v}$ to yield $d_v$-dimensional output values. The model then concatenates all the outputs and projects them back to a $d$-dimensional representation, as illustrated in the following equations:
\begin{align}
\label{Eq_1112}
\operatorname{LGS-MHSA}(\bm{Q}\text{, } \bm{K}\text{, }\bm{V}) &= \operatorname{Concat}(\text{H}_1,\ldots\text{, }\text{H}_h)W^O\text{,} \\
\operatorname{where} \text{H}_i &= \operatorname{LGA}(\bm{Q}_{d_k}\text{, } \bm{K}_{d_k}\text{, } \bm{V}_{d_v})\text{,}
\end{align}
where $h$ is the number of heads in model, $\bm{Q}\text{, }\bm{K}\text{, }\bm{V} \in \mathbb{R}^{T\times d}$, $\bm{Q}_{d_k}\text{, }\bm{K}_{d_k}\text{, }\bm{V}_{d_v} \in \mathbb{R}^{T\times d_k}$, $W^O\in\ \mathbb{R}^{hd_v\times d}$. In this work, we use $h=8$, and thus $d_k=d_k=d_v=d/h=8$.

Furthermore, it is noteworthy that sparse attention is implemented only on the encoder side. This is because the encoder processes the original input sequence, which accounts for most of the computation, whereas the decoder operates on the summary sequence, which is typically much shorter than the original input. In addition, the computational complexity of the local attention pattern is $O(n\times w)$, scaling linearly with the length $ n $ of the input frame sequence. Although global attention is also adopted in this work, since the number of global tokens is relatively small and independent of $ n $, the overall computational complexity of LGA remains $O(n)$.

\subsection{Training}
During training, we mainly focus on the teacher forcing method used by the decoder, which is because the decoder uses it to process the summary sequence. At each time step, instead of directly using its previous prediction, the decoder considers taking the target output sequence, i.e., ground truth, from the dataset as its input. This approach allows the model to effectively and accurately update its parameters to be learned based on the correct error backpropagation, which is beneficial to train a good model in a fast convergence way, so as to make the trained model generate the desired summary sequence. Teacher forcing helps the model learn how to generate the next token, while the masking mechanism ensures that the model does not ``cheat" by peeking at future tokens during training. The combination of the two techniques enables the model to more effectively learn how to generate accurate target sequences. Additionally, implementing the LGA sparse mechanism is non-trivial, since the local attention requires a form of banded matrix multiplication. 
To address this, we adopt a customized CUDA kernel, as used in \cite{beltagy2020longformer}, for efficient implementation.

We employ the simple and computationally efficient binary cross-entropy (BCE) loss to train model. We treated the label of each video frame as a binary classification problem to indicate whether a frame is selected as a keyframe or not. The loss function is formally denoted as follows:
\begin{equation}
\label{Eq_13}
    \begin{split}
     {\mathcal{L}_{BCE}(y_n\text{, } p_n)=}  \quad \quad \quad \quad \quad \quad \quad  \quad \quad \quad \quad \quad \quad \quad \\ {-\frac{1}{T}\sum_{n=1}^{L \times T}\left[y_nlog(p_n)+\left(1-y_n\right)log\left(1-p_n\right)\right] \text{,}} 
    \end{split}
\end{equation}
where $T$ represents the overall number of video frames, and $y_n$ denotes the ground-truth label of the $n$-$th$ frame, obtained by converting importance scores into binary labels (0 and 1) at the shot level. The importance score can be interpreted as the probability $p_n$ of a keyframe being selected.  The training procedure of our FullTransNet is summarized in Algorithm \ref{algorithm1}.
\begin{algorithm}[htbp]
\caption{The train procedure of FullTransNet}
\label{algorithm1}
\SetAlgoLined
\LinesNumberedHidden
\KwIn{Original videos $\bm{X}_n$, Frame-level importance scores $\bm{y}_n$.}
\KwOut{Video summaries $\bm{p}_n$ and all parameters $\theta$ in FullTransNet.}
Initialize $\theta$ of FullTransNet using Xavier. \\
\For{split $\leftarrow$ 1 \KwTo 5}
{
  \For{epoch $\leftarrow$ 1 \KwTo E}
  {\% $E$ indicates epoch, $E=300$ in our work.\ \\
    \For{$\bm{X} \in \{\bm{X}_n\}$, $\bm{y} \in \{\bm{y}_n\}$}
    {
      \% key-frames summary \;
      $\bm{s}$ $\leftarrow$ Knapsack \cite{song2015tvsum} $(\text{KTS \cite{potapov2014category}}(\bm{X}), \bm{y})$ \;
      $\bm{S}$ $\leftarrow \bm{X} [ \,\text{if}\,  \bm{s} \, \text{is\, True} ] $ \;
      $\bm{E}_X \leftarrow embedding(\bm{X})+ \bm{P}_{enc}(\bm{X})$  \;
      $\bm{E}_S \leftarrow embedding(\bm{S})+ \bm{P}_{dec}(\bm{S})$  \;
      $\bm{Y}_{enc}^{N-1} \leftarrow encoder(\bm{E}_X) $ \;
      $\bm{Y}_{enc}^{N} \leftarrow encoder(\bm{Y}_{enc}^{N-1}) $ \;
      $\bm{Y}_{dec}^{N-1} \leftarrow decoder(\bm{E}_S, \bm{Y}_{enc}^{N}) $ \;
      $\bm{Y}_{dec}^{N} \leftarrow decoder(\bm{E}_S, \bm{Y}_{dec}^{N-1}) $ \;
      $\bm{p} \leftarrow softmax(Linear(\bm{Y}_{dec}^{N}))$ \;
      $\bm{p}_n \leftarrow ScoretoKeyFrame(\bm{p})$ \;
      $\mathcal{L} \leftarrow \mathcal{L}_{BCE}(\bm{y}_n, \bm{p}_n)$ \; 
      \% i.e., loss in  Eq. (\ref{Eq_13}) \;
      $\theta \leftarrow \theta - \alpha   \frac{\partial \mathcal{L}}{\partial \theta } $ \;
    }
  }
}
\end{algorithm}

\section{Experiments and Results}
\label{sec:exp-results}
\subsection{Datasets}

Two public benchmark datasets: SumMe \cite{gygli2014creating} and TVSum \cite{song2015tvsum} for video summarization are used to evaluate the proposed method. SumMe consists of 25 user videos ranging from 1 to 6 minutes in length and provides multiple user-annotated summaries (by 15-18 different users) for each video in the form of key shots. The dataset covers multiple events from both first-person and third-person cameras, such as holidays, cooking, and sports. TVSum contains 50 videos with 10 categories (5 videos per category), which vary from 2 to 10 minutes in length and are annotated by 20 users in the form of frame-level importance scores. Similar to SumMe, the dataset also covers various genres such as beekeeping, making sandwiches, and grooming an animal. Since our method is based on supervised learning, more annotated data is more beneficial to the training. Thus, two other annotated datasets with keyframes: OVP \cite{de2011vsumm} and YouTube \cite{de2011vsumm} are used as augmented training datasets. OVP has 50 videos with various genres varying from 1 to 10 minutes in length, while YouTube consists of 39 videos with multiple visual styles excluding cartoons, whose lengths are from 1 to 4 minutes. The descriptions of four datasets are shown in Table \ref{table1}.

Following previous works \cite{Zhang2016video}, \cite{li2022joint}, \cite{mahasseni2017unsupervised}, \cite{ji2019video},  \cite{ji2020deep}, we apply three commonly used settings, i.e., canonical, augmented, and transfer, to train and evaluate our model. 
Concretely, in the canonical setting, with respect to SumMe and TVSum, training and testing sets are from the same dataset between them. Each dataset is randomly divided into 5 disjoint splits, with 80\% of the data used for training and the remaining 20\% used for evaluation.
In the augmented setting, YouTube and OVP are added to the training set, and the testing set remains the same. As for the transfer setting, YouTube, OVP, and TVSum (SumMe) are used as the training set, and SumMe (TVSum) is used as the testing set.

Since our model requires ground-truth key shots, for the TVSum dataset, which only provides frame-level importance scores, we follow \cite{Zhang2016video} to convert the frame-level importance scores into key shots.
In addition, following common practice \cite{Zhang2016video}, \cite{li2022joint}, \cite{mahasseni2017unsupervised}, \cite{ji2019video}, \cite{fajtl2019summarizing}, we use kernel-based temporal segmentation (KTS) \cite{potapov2014category} to detect shot boundaries and apply the knapsack algorithm \cite{song2015tvsum} to ensure that the generated video summary does not exceed 15\% of the original video length.
Although there are more advanced shot boundary detection algorithms, such as \cite{Idan2021SBD}, we did not adopt them in our work to ensure a fair comparison with other methods.
\begin{table*}[htbp]
\centering
\caption{The Descriptions of Datasets Used in the Experiments. ``\# Frames'' indicates the number of frames in all shots.}
\resizebox{1.0\textwidth}{!}{
\begin{tabular}{l|c|c|c|c|c}
% \begin{tabular}{l|c|c|c|c|ccc}
\toprule
% \hline
% \hline
Dataset & \# Videos & Duration (min)   & Genres/Topics & Annotations & \# Frames (Min,Max,Avg)
\\ 
% \hline
% \hline
\midrule
SumMe \cite{gygli2014creating} & 25    & 1--6     & Holidays and Sports & Key shots &{ 7},\quad2133,\quad146 \\
TVSum \cite{song2015tvsum} & 50    & 2--10    & News Coverage, and Bee Activities & Frame-level important scores &10,\quad{ 997},\quad148 \\
OVP \cite{de2011vsumm}    & 50    & 1--10    & Cartoons, Commercials, and Home Videos  & Keyframes &29,\quad1289,\quad213 \\
YouTube \cite{de2011vsumm} & 39    & 1--4     & Documentary, Educational, and Lecture & Keyframes &15,\quad3780,\quad221 \\ 
\bottomrule
\end{tabular}
}
\label{table1}
\end{table*}

\subsection{Evaluation Metrics}

For a fair comparison with other previous works  \cite{Zhang2016video}, \cite{li2022joint}, \cite{ji2019video}, \cite{fajtl2019summarizing},\cite{sen2024multi} we adopt F-Measure (a.k.a. F-score) as the evaluation metric. Given a video $V$, $V_{gt}$ and $V_{gs}$ represent its ground-truth summary and generated summary, respectively. Precision($P$) and recall ($R$) are then calculated based on the length of temporal overlap between $V_{gt}$ and $V_{gs}$ as follows:
\begin{align} 
\label{Eq_14}
P=\frac{|V_{gt} \cap V_{gs}|}{| V_{gs}|}\text{, } R=\frac{|V_{gt} \cap V_{gs}|}{| V_{gt}|}\text{,}
\end{align}
where $V_{gt} \cap V_{gs}$ denotes temporal overlap between them, and \iffalse $\lvert \bullet \lvert$ \fi $| \cdot |$ indicates the length of temporal duration. The harmonic mean F-Measure  is then  computed with $P$ and $R$ as follows:
\begin{align} 
\label{Eq_15}
F=\frac{2PR}{P+R}\times100\%.
\end{align}
A higher F-Measure means more temporal overlaps between the generated summary and ground-truth summary while keeping less redundancy.

\subsection{Experimental settings}
\label{subsec:Exp-setting}

We train our FullTransNet and its various ablation variants on an NVIDIA 3090 graphics card with 24GB memory. We implement our approach using Python 3.10 and PyTorch 2.0 \cite{Paszke2017AutomaticDI}. We train all models using Adam \cite{kingma2014adam} optimizer for 300 epochs from scratch on corresponding datasets in three different settings, with a batch size of 1, learning rate of 1$e$--3, momentum of 0.9, and weight decay of 1$e$--4. Training one model takes 5--8 hours. We set the size of the sliding window to 17, that is, each token attends to 8 tokens on both sides of itself. Empirically, the tokens in this attention span have proved to be most relevant. Additionally, we set the feature dimension of the token as 64. 
This is done to achieve a trade-off between computation efficiency and model complexity. 
In this task, since the maximum length of videos in all datasets is 1,513 frames, to accommodate all video frames and compute weight matrices, we set the sequence length to 1,536. As for these sequences whose length is less than the maximum value, we pad ``0’’ into the corresponding positions in these sequences to keep the length of each video sequence identical.

Besides, following the convention in \cite{Zhang2016video}, \cite{li2022joint}, \cite{mahasseni2017unsupervised}, \cite{ji2019video}, all videos are downsampled from 30 fps to 2 fps to reduce redundancy. The output of the pool5 layer of GoogLeNet \cite{szegedy2015going}, pre-trained on ImageNet \cite{olga2015imagenet}, is used as the feature descriptor for each video frame. Five-fold cross-validation is performed for each setting to compute the average performance.

\subsection{Comparisons with Existing Methods}

In this section, we compare the proposed FullTransNet with several existing methods on SumMe and TVSum in terms of F-Measure. Among these representative works, vsLSTM \cite{Zhang2016video}, dppLSTM \cite{Zhang2016video}, H-RNN \cite{zhao2017hierarchical}, HSA-RNN \cite{zhao2018hsarnn}, TTH-RNN \cite{zhao2020tthrnn}, and SUM-FCN \cite{rochan2018video} primarily adopt RNNs or CNNs as core building techniques. In contrast, A-AVS \cite{ji2019video}, M-AVS \cite{ji2019video}, vsLSTM+Att \cite{casas2018video}, dppLSTM+Att \cite{casas2018video}, SUM-GAN-AAE \cite{apostolidis2020unsupervised}, DASP \cite{ji2020deep}, VASNet \cite{fajtl2019summarizing}, SUM-GDA$_{\text{sup}}$ \cite{li2021exploring}, H-MAN \cite{liu2019learning}, DMASum \cite{wang2020query}, VJMHT \cite{li2022joint}, and STVT \cite{hsu2023video}  use attention mechanisms as key building components. Notably, methods such as \cite{li2022joint}, \cite{li2021exploring}, \cite{fajtl2019summarizing}, \cite{liu2019learning}, \cite{hsu2023video}, and \cite{wang2020query} utilize transformer-based attention, while the others employ non-transformer attention mechanisms.

The quantitative comparisons of results are shown in Table \ref{table2}. It can be seen that the summarization performance of models with attention consistently surpasses that of other models without attention under almost all the three commonly used settings, i.e., canonical, augmented, and transfer settings. It shows that the attention mechanism is capable of modeling long-range dependency in sequence data of video, which is of central importance for the performance-boosting of video summarization tasks. Furthermore, there is a similar observation between non-transformer and transformer attention, in which transformer attention typically outperforms non-transformer on the summarization performance. We believe that the reason may be because the self-attention mechanism \cite{vaswani2017attention} without recurrence and convolution operations can learn better representation in sequences relative to other attentions. More concretely, one can see that our FullTransNet outperforms almost all the existing approaches on both two datasets under all three settings except for DASP \cite{ji2020deep} and DMASum \cite{wang2020query}. DASP \cite{ji2020deep} slightly improves by 0.4\% solely on TVSum under the augmented settings, while our method significantly surpasses DASP \cite{ji2020deep} on SumMe by 8.9\% and 7.6\% under both the canonical and augmented settings, respectively, even on TVSum by 0.3\% under the augmented settings. This clearly shows that our methods can achieve a well-balanced performance on the two datasets, verifying its effectiveness and advancement, while DASP \cite{ji2020deep} may be a method for specific datasets. DMASum \cite{wang2020query} outperforms our method by 1.2\% and 1.3\% on both SumMe and TVSum only under the transfer settings. We conjecture that FullTransNet may have overly learned these different types of video data due to its relatively high network capacity which leads to a low generalization ability.
\begin{table*}[htbp]
\centering 
\small
\caption{The F-Measure (\%) of different methods under three settings. The best and the second-best results are in \textbf{blod} and \underline{underlined}, respectively. $\star$ indicates that the results are reproduced using publicly available code under the same GoogLeNet feature descriptor and down-sampling strategy settings as other methods for a fair comparison. $\dagger$ denotes that since we do not find the corresponding code, here the listed results are from the published paper, whose method uses the VGG-16 feature descriptor but adopts the same down-sampling strategy as all other methods.}
\resizebox{1\linewidth}{!}{
% \begin{tabular}{l|c|ccc|ccc}
\begin{tabular}{p{3cm} | p{3cm}<{\centering} | p{1.5cm}<{\centering} p{1.5cm}<{\centering} p{1.5cm}<{\centering}| p{1.5cm}<{\centering} p{1.5cm}<{\centering} p{1.5cm}<{\centering}}
\toprule
% \hline
\multirow{2}{*}[-0.8ex]{Method} &\multirow{2}{*}[-0.8ex]{Main Techniques} & \multicolumn{3}{c|}{SumMe}       & \multicolumn{3}{c}{TVSum}       
\\
\cmidrule{3-8}
           &      &  Canonical & Augmented & Transfer & Canonical & Augmented& Transfer
\\ 
\midrule
vsLSTM \cite{Zhang2016video}   &  \multirow{6}{*}{RNN or CNN} &    37.6     &    41.6     &    40.7     &   54.2       & 57.9     & 56.9    
\\
dppLSTM \cite{Zhang2016video}     &     &  39.6    &   42.9    &   41.8    & 54.7     & 59.6     & 58.7    
\\
H-RNN \cite{zhao2017hierarchical}         &  & 44.3  &     --    &  --   &   62.1     &  --   &    --     
\\
HSA-RNN$^\dagger$ \cite{zhao2018hsarnn}         &  & 44.1  &     --    &  --   &   59.8     &  --   &    --     
\\
TTH-RNN \cite{zhao2020tthrnn}         &  & 45.0  &     --    &  --   &   62.3     &  --   &    --     
\\
SUM-FCN  \cite{rochan2018video}         &  & 47.5  &     51.1    &  44.1   &   56.8     &  59.2   &    58.2     
\\
\midrule
A-AVS \cite{ji2019video}     & \multirow{6}{*}{\shortstack{Non-Transformer\\Attention}} & 43.9   &    44.6     &  --       & 59.4     &  60.8      & --       
\\
M-AVS \cite{ji2019video}         &   &   44.4   &    46.1    &   --      &   61.0     & 59.1     & --       
\\
vsLSTM+Att \cite{casas2018video}  &  & 43.2  &     --    &  --   &   63.1     &  --   &    --   
\\
dppLSTM+Att \cite{casas2018video}  &  & 43.8  &     --    &  --   &   53.9     &  --   &    --   
\\
SUM-GAN-AAE \cite{apostolidis2020unsupervised} &  & 48.9  &     --    &  --   &   58.3     &  --   &    --   
\\
DASP   \cite{ji2020deep}           &   & 45.5   &    47.0    &    --       & \underline{63.6}     &  \textbf{64.5}   & --         
\\
\midrule
VASNet \cite{fajtl2019summarizing}       &  \multirow{7}{*}{Transformer Attention}  &  49.7    &   51.1   &    --       & 61.4     & 62.3     & --      
\\
SUM-GDA$_{\text{sup}}$ \cite{li2021exploring}     &   &   52.8   &    \underline{54.4}      & 46.9    & 58.9  &    60.1     & 59.0       
\\
H-MAN  \cite{liu2019learning}          &   &   51.8      & 52.5     & 48.1  &  50.4    &   61.0   &    59.5
\\
DMASum  \cite{wang2020query}          &   &  \underline{54.3}      & 54.1      & \textbf{52.2}    & 61.4      & 61.2     & \textbf{60.5} 
\\
VJMHT   \cite{li2022joint}          &   & 50.6   &   51.7    & 46.4    & 60.9   &   61.9    &  58.9        
\\
STVT$^\star$   \cite{hsu2023video}          &   & 50.8   &   --    & --   & 61.7   &   --    &  --     
\\
FullTransNet (ours)  & &  \textbf{54.4}   &   \textbf{54.6}    &  \underline{51.0}     & \textbf{63.9}     &  \underline{64.1}    & \underline{59.2}   
\\
\bottomrule
\end{tabular}
}
\label{table2}
\end{table*}

\subsection{Ablation Study}

In this section, we conduct an extensive ablation study on both two datasets under the canonical setting, to thoroughly evaluate the impact of each component involved in FullTransNet on the performance. Specifically, we study and analyze the impact of local-global sparse attention (LGA) module, the number of encoder layers with LGA, the dimension of hidden layer in FFN, the dimension of frame embedding, and the number of heads in LGS-MHSA, window size (WS) in local attention (LA), and the number of global tokens in global attention (GA). All settings are the same as those mentioned in Subsection \ref{subsec:Exp-setting} unless specified otherwise. \iffalse The quantitative results are shown in Table \ref{table3} and in Table \ref{table4}. \fi

\subsubsection{Impact of Local-Global Sparse Attention Module}

The main purpose of conducting this ablation is to test the impact of different sparse attention patterns, including LA, GA, and LGA, and to compare sparse attention with full attention (FA) on the summarization performance. The results are shown in Table \ref{table3}. It can be seen that FA generally yields better results compared to LA and GA, but it has a higher computational complexity, leading to significant computational and memory overhead. Additionally, LGA, a combination of LA and GA, achieves the best F-Measure by 54.4\% and 63.9\% on SumMe and TVSum datasets, respectively, and compared to FA, it has a lower computation cost, a shorter inference time, and a less memory footprint. The superior performance indicates that our LGA module has a considerable advantage in modeling long-term dependencies and saving computing and memory costs. This is beneficial for improving the performance in long sequence modeling tasks such as video summarization. The observation and finding are also consistent with that in BigBird \cite{zaheer2020bigbird} used for text summarization. It is worth noting that the FLOPs, runtime, and memory are calculated with the 18-$th$ video of SumMe and the 45-$th$ video of TVSum, respectively. The former consists of 149 frames, with 44 frames executing global attention, while the latter contains 166 frames, with 51 frames executing global attention.
\begin{table*}[htbp]
\caption{Comparisons on F-Measure (\%) of different attention patterns on SumMe and TVSum. FA represents full attention. LA and GA denote local attention and global attention, respectively. LGA means a combination of local attention and global attention. Note that all attention patterns only be adopted at the encoder side; all evaluations are performed under the canonical setting. Other parameters are the same as those of FullTransNet in Table \ref{table4}.}
\centering
\resizebox{1.0\linewidth}{!}{
% \begin{tabular}{c|cccc|c|c|c|c|c}
\begin{tabular}{p{1.5cm}|p{0.8cm}<{\centering}  p{0.8cm}<{\centering}  p{0.8cm}<{\centering}  p{0.8cm}<{\centering} |p{1.5cm}<{\centering} |p{1.5cm}<{\centering} |p{1.5cm}<{\centering}| p{1.5cm}<{\centering}|p{1.5cm}<{\centering}}
\toprule
\multirow{2}{*}[-0.8ex]{Dataset} &\multicolumn{4}{c|}{Attention Pattern}& \multirow{2}{*}[-0.8ex]{F-Measure}& \multirow{2}{*}[-0.8ex]{\shortstack{Params (M)} }&\multirow{2}{*}[-0.8ex]{\shortstack{FLOPs (G)}}&\multirow{2}{*}[-0.8ex]{\shortstack{Runtime (s)}}&\multirow{2}{*}[-0.8ex]{\shortstack{Memory (GB)}}
\\
\cmidrule{2-5}
&FA & LA & GA & LGA & &&&&
\\
\midrule
\multirow{4}{*}{SumMe}& $\checkmark$ & --& -- & --& 52.67 & 3.613&2.785 &1.345&0.171
\\
&--&  $\checkmark$ & --& --&  53.86     &  3.525   & 2.583 &0.005&0.030
\\
&--& -- & $\checkmark$ & --&  51.91    & 3.525    &2.639 &0.007&0.031
\\
&--&-- & --& $\checkmark$ & 54.40 & 3.538 &2.672 &0.117&0.038
\\
\midrule
\multirow{4}{*}{TVSum} & $\checkmark$ & --& -- & -- &  62.75     & 3.613    & 2.899 &2.338&0.171
\\
&--&  $\checkmark$ & --& --& 62.12 &3.525 &2.697 &0.019&0.030
\\
&--& -- & $\checkmark$ & --&  61.82    & 3.525    & 2.753&0.050&0.032
\\
&--&-- & --& $\checkmark$ & 63.94 &3.538  & 2.786&0.179&0.039
\\
\bottomrule
\end{tabular}
}
\label{table3}
\end{table*}

\subsubsection{Impact of the Number of Encoder Layers with LGA}

Empirically, a greater number of encoder layers generally means the model has stronger representation ability because more layers in the encoder can capture more complex and higher-level features, and more layers in the decoder allow the model to consider more contextual information, leading to more accurate results. However, a bigger model requires more computational cost, and could potentially result in over-fitting issues. Thus, we conduct this ablation to demonstrate how different numbers of layers affect the model performance. In this ablation, the number of layers in the encoder and decoder is set to 2, 4, 6, and 8, respectively. From Table \ref{table4} rows (A), it can be observed that as the number of layers in the encoder and decoder increases, the performance gradually improves on SumMe, and the best result can be achieved by the FullTransNet with 6 layers in both encoder and decoder. But under an 8-layer setting, the performance drops significantly, which implies over-fitting. As for the number of layers, there is a similar observation on TVSum, with only an exception where the model with 2 layers slightly improves by 0.01\% compared to the model with 4 layers. We believe that this reason should be attributed to the dataset itself. The distribution of data on SumMe may be more uniform, which is beneficial for model training. Therefore, the encoder with 6 layers is used throughout all the experiments.
\begin{table*}[htbp]
\caption{Performance comparisons on F-Measure (\%) of FullTransNet architecture and its ablation variants on SumMe and TVSum. Default values are identical to those of the FullTransNet model. All evaluations are performed under the canonical setting.}
\centering 
\resizebox{1\linewidth}{!}{
% \begin{tabular}{c|cccccc|c|c|cc}
\begin{tabular}{p{1.5cm}<{\centering} | p{0.6cm}<{\centering} p{0.6cm}<{\centering} p{0.6cm}<{\centering} p{0.6cm}<{\centering} p{0.6cm}<{\centering} p{0.6cm}<{\centering} |p{1.5cm}<{\centering} |p{2cm}<{\centering}|p{1cm}<{\centering} p{1cm}<{\centering}}
\toprule
                     & $N$ & $d$ & $d_{ff}$  & $h$ & $d_k$  & $d_v$  & WS in LA&\# Tokens in GA & SumMe    & TVSum 
                     \\ 
\midrule
\multirow{3}{*}{(A)} & 2 &        &      &   &     &     &       &       & 50.6         & 63.1 
\\
                     & 4 &        &      &   &     &     &         &     & 52.9           & 63.0 
                     \\
                      & 8 &        &      &   &     &     &         &     & 51.3          & 62.7 
                     \\ 
\midrule
\multirow{4}{*}{(B)} &   & 32     &      &   & 4   & 4   &         &     & 51.1          & 63.0 
\\
                      &   & 256    &      &   & 32 & 32  &           &   & 49.7          & 59.6
                     \\
                     &   & 512    &      &   & 64  & 64  &          &    & 44.8          & 57.8 
                     \\
                     &   & 1024   &      &   & 128 & 128 &         &     & 46.6          & 56.9 
                     \\ 
\midrule
\multirow{2}{*}{(C)} &   &        & 512  &   &     &     &         &     & 52.4          & 64.0 
\\
                     &   &        & 1024 &   &     &     &         &     & 52.2          & 63.8 
                     \\
\midrule  
\multirow{2}{*}{(D)} &   &        &      & 1 & 64  & 64  &        &      & 45.7          & 60.1
\\
                     &   &        &      & 4 & 16  & 8   &        &      & 48.9          & 62.8
                     \\

\midrule
\multirow{3}{*}{(E)} &   &        &      &   &     &     & 9       &     &49.8           &   63.9    
\\
                     &   &        &      &   &     &     & 65       &    & 53.3          & 63.7      
                     \\
                     &   &        &      &   &     &     & 129      &    & 52.2           & 63.3      
                     \\ 
\midrule
\multirow{2}{*}{(F)} &   &        &      &   &     &     &       &  1   &52.7           &   62.8    
\\
                     &   &        &      &   &     &     &       & 2   & 53.0          & 63.1     
                     \\
\midrule
FullTransNet  & 6 & 64     & 2048 & 8 & 8   & 8   & 17     &3      & \textbf{54.4} & \textbf{63.9} 
\\ 
\bottomrule
\end{tabular}
}
\label{table4}
\end{table*}

\subsubsection{Impact of the Dimension of Frame Embedding}

The dimension of frame embeddings is important to the performance of the model. In this section, we conduct this ablation on the dimension of token embeddings to evaluate the performance of the model with different dimension settings. The results generated using five different dimensions: 32, 64, 256, 512, and 1,024, are shown in Table \ref{table4} rows (B). It can be seen that when this dimension is set as 64, the performance reaches the best on the two datasets, while after that, the performance is nearly gradually decreased as the dimension increases. We believe that over-fitting could occur in those cases. Thus, we set the dimension as 64 and use it for all the experiments.

\subsubsection{Impact of the Dimension of Hidden Layer in FFN}

In the ablation study, our main purpose is to explore how different dimensions ($d_{ff}$) of the middle hidden layer in the feedforward network affect the performance of the network. We vary the number of neurons in the hidden layer to measure the performance. By comparing the performance of models with different dimensions of the hidden layer, we aim to achieve the best performance under a specific value of $d_{ff}$. The results are shown in Table \ref{table4} rows (C). One can see that the performance on the two datasets is consistently decreasing when the dimensionality $d_{ff}$ changes from 512 to 1,024, while the performance can be consistently improved when the dimensionality $d_{ff}$ goes from 1,024 to 2,048. We believe the reason is that when the $d_{ff}$ is fixed to 1,024, other parameters may not be the optimal hyperparameters, thus making the model trapped in a local optimum; In contrary, when $d_{ff}$ is set to 2,048, combined with other parameters, which makes it possible to learn an optimal model with good generalization ability. Therefore, we set the dimension of the hidden layer in FFN to 2,048 in FullTransNet.

\subsubsection{Impact of the Number of Heads in LGS-MHSA}

In this section, we will explore the impact of the number of heads in the local-global sparse multi-head self-attention (LGS-MHSA) mechanism on performance. The results are shown in Table \ref{table4} rows (D). 
As we can see, the performance on both datasets gradually improves as the number of heads increases, with the best results achieved using 8 heads.
This reason may be that more heads in LGS-MHSA attend to more information at different positions from different representation subspaces, which could enhance the representation ability of the model. The visualization of the information attended by 8 heads in each layer at different positions is detailed in Subsection \ref{visual}. The qualitative results show how the query token focuses on the information at different positions.

\subsubsection{Impact of Window Size in Local Attention}

To show the importance of window size in local attention, we investigate the impact of window size using different configurations of 9, 17, 65, and 129, and report the experimental results in Table \ref{table4} rows (E). It can be seen that using a window size of 17 achieves the best results on both SumMe and TVSum datasets; when using window sizes of 65 and 129, the performance seems to be gradually decreasing, while using a window size of 9, the best performance can be achieved only on TVSum. The observation demonstrates that modeling short-term dependencies is more crucial than modeling long-term dependencies for video summarization. The reason may be that if the window size is too large, the attention may be spread to many irrelevant tokens, distracting from the key local information. The distracting attention can prevent the model from effectively capturing the information of important tokens, resulting in decreased accuracy. This assumption can be proved by the evidence from Table \ref{table2}. One can see that our LA achieves results on par with or better than FA on both datasets. This is consistent with the common sense that, for video summarization tasks, a keyframe is more relevant to the frames in its neighborhood than to all other frames in the whole sequence, especially for extractive summarization tasks. Thus, for all the experiments, we use the fixed configuration of $\operatorname{WS}=9$. 

\begin{figure*}[htbp]
\centering
\includegraphics[width=1\linewidth]{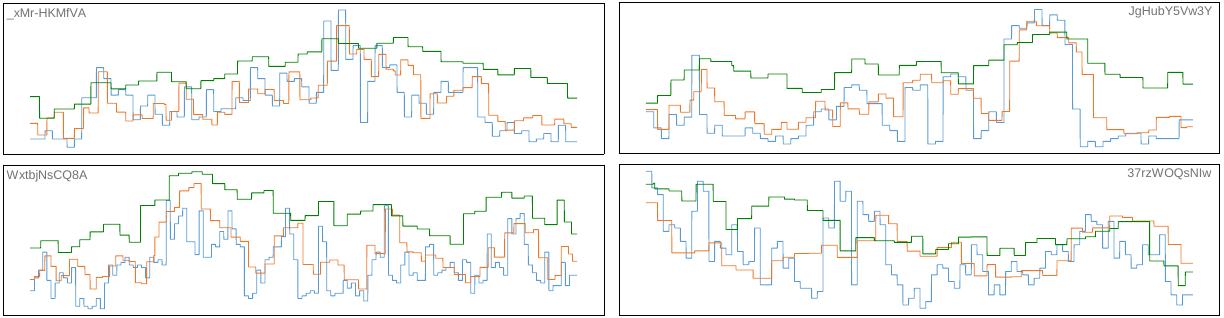}
\caption{Comparison of the predicted importance scores generated by FullTransNet with LGA and FA, as well as the ground-truth importance scores, on four videos from the TVSum dataset. Each subplot corresponds to one video and displays three score lines across all video frames, where the \textcolor{orange}{orange} line, the \textcolor{green}{green} line, and the \textcolor{blue}{blue} line represent the scores generated by FullTransNet with LGA, FullTransNet with FA, and ground-truth scores, respectively. The name of each video is placed at the top-left or top-right corner of its respective subplot.}
\label{fig3}
\end{figure*}

\begin{figure*}[htbp]
\centering
\includegraphics[width=1\linewidth]{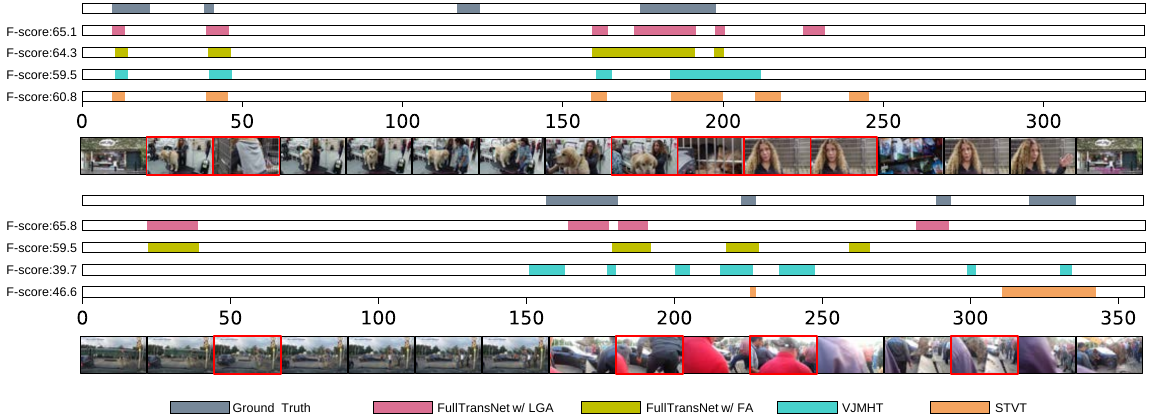}
\caption{Qualitative comparisons of different summarization methods, including ground truth, FullTransNet with LGA and FA, VJMHT \cite{li2022joint}, and the recent STVT \cite{hsu2023video}, on the 11-$th$ video from TVSum (top) and the 6-$th$ video from SumMe (bottom). The horizontal axis represents the frame indices, while different colors in the bar indicate summaries generated by the respective methods. Frames selected as summary frames by LGA are highlighted with red boxes, sampled every 20 frames from the videos.}
\label{fig4}
\end{figure*}

\subsubsection{Impact of the Number of Tokens in Global Attention}
The impact of the number of tokens in global attention is shown in Table \ref{table4} rows (F). We can see that as the number of tokens increases, the performance is also improved. However, this does not mean that a higher number of tokens has a better performance. The evidence from Table \ref{table2} can support the viewpoint. As mentioned in Section \ref{sec:method}, we only take the first, the middle, and the last frame of a shot as the global tokens. 
Here, the token counts of 1, 2, and 3 correspond to using only the first frame, both the first and middle frames, and all three frames, respectively.

\subsection{Visualization}
\subsubsection{Predicted Scores of FullTransNet and Its Variants}

In this section, we will give a comprehensive qualitative evaluation to intuitively demonstrate the performance of our method and its variants on the 35-$th$, 44-$th$, 36-$th$, and 19-$th$ videos from the TVSum dataset. The ground truth and predicted scores by FullTransNet are illustrated in Fig. \ref{fig3}. The \textcolor{blue}{blue} lines depict the ground truth scores and the \textcolor{orange}{orange} lines show the generated scores by FullTransNet with LGA. One can see that the predicted scores fit their corresponding ground-truth values very well on the four videos. Moreover, we also visualize the predicted important scores by FullTransNet without LGA, i.e., with FA, indicated by the \textcolor{green}{green} lines. It can be seen that compared to the \textcolor{green}{green} lines, the \textcolor{orange}{orange} lines are closer to the ground-truth lines. 
This demonstrates that LGA is effective in measuring the relative importance between frames and selecting the most relevant frames as keyframes.
Notably, the visual results are generated from four videos spanning different categories, demonstrating that our FullTransNet model consistently and effectively aligns with the ground-truth scores across diverse video types. This highlights the model's robustness and adaptability to various types of content.

\subsubsection{Summary Segments Generated by Different Methods}

To further understand the summarization results, we compare our proposed FullTransNet with FA and LGA with recently published VJMHT \cite{li2022joint} and STVT   \cite{hsu2023video} on the 11-$th$ video (about animal grooming) from TVSum (top) and the 6-$th$ video (about a car across railway) from SumMe (bottom). The qualitative results in terms of summary segments generated by different methods are shown in Fig. \ref{fig4}, where the results using STVT are achieved by down-sampling to 2 fps for a fair comparison. One can see that our proposed FullTransNet with FA and LGA has high overlaps with the ground truth. Although FA and LGA generate similar summary segments, the summary segment generated by LGA seems to be close to the ground truth. 
Additionally, both VJMHT and STVT achieve performance that is nearly comparable to our FullTransNet with LGA on the 11-$th$ video of TVSum. However, on the 6-$th$ video of SumMe, VJMHT produces fewer segments that overlap with the ground truth, despite effectively distinguishing shots. On the other hand, STVT generates more concentrated summaries, which deviate significantly from the ground truth.

As for the reasons, aside from the dataset itself, we believe that using LGA instead of FA prevents the model from applying weighted averaging to all tokens. This allows the model to better distinguish between different shots and select more representative key shots. Moreover, as shown in Table \ref{table3}, LGA exhibits lower computational complexity compared to FA, demonstrating its superiority for video summarization tasks.

\subsubsection{Attention Maps}
\label{visual}

\begin{figure*}[htbp]
\centering
\includegraphics[width=1\linewidth]{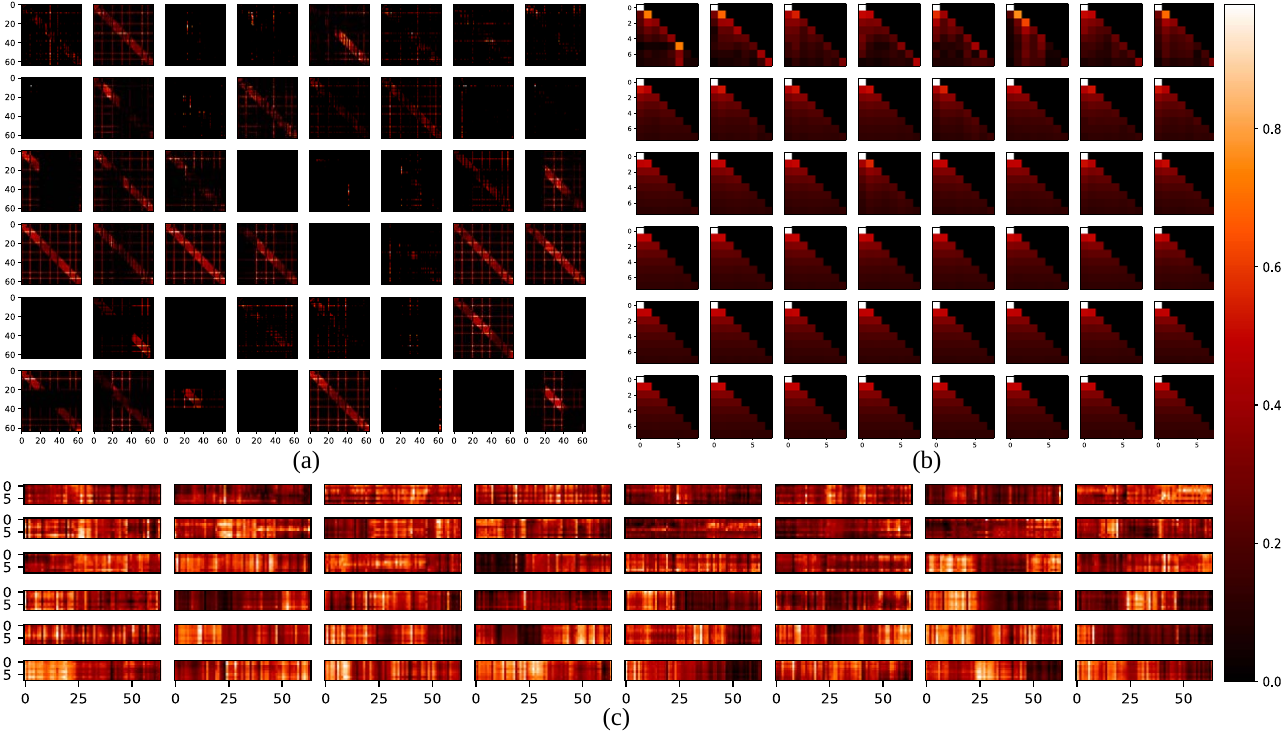}
\caption{Visualization of (a) LGA in encoder, (b) masked in decoder, and (c) encoder-decoder attention map for the 12-$th$ video of SumMe,  all with 6 layers and 8 heads, which is from the first epoch during training.}
\label{fig5}
\end{figure*}

\begin{figure}[htbp]
\centering
\includegraphics[width=1\linewidth]{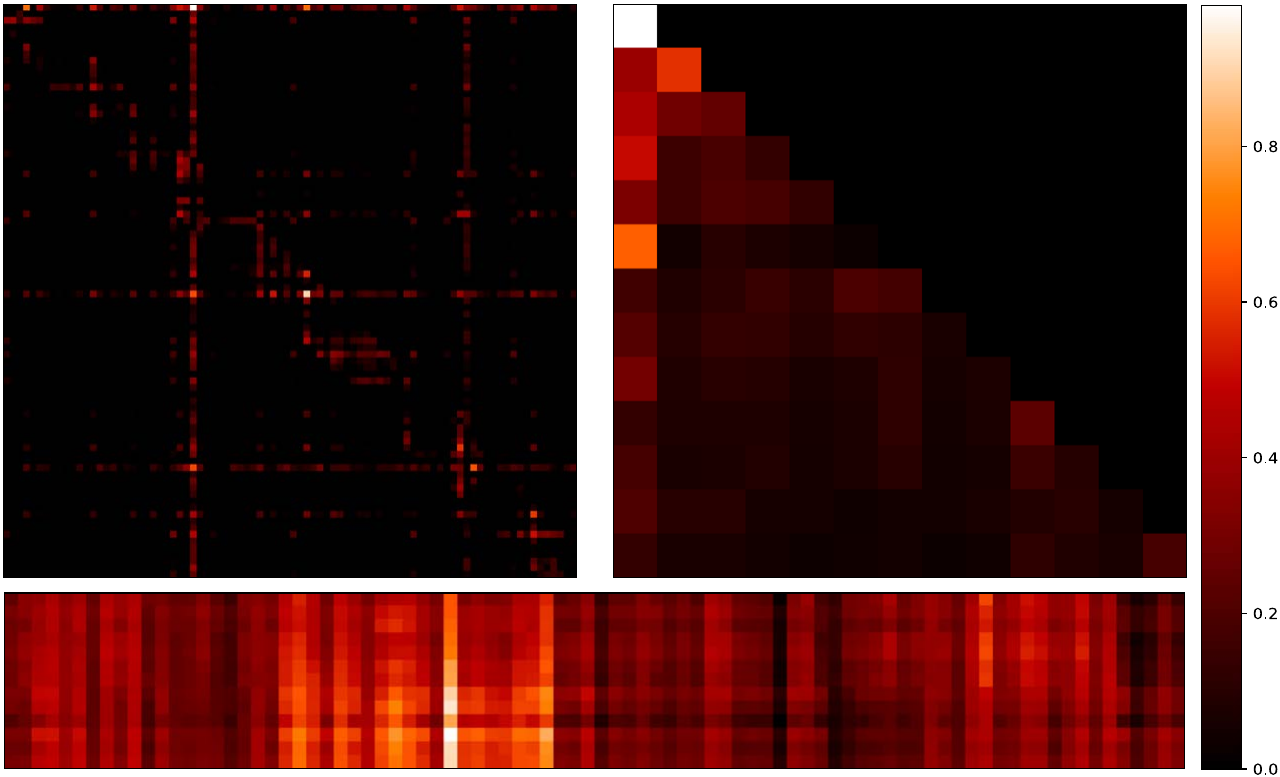}
\caption{Visualization of LGA (top left), masked (top right), and encoder-decoder (bottom) attention map with one head for the 8-$th$ video of SumMe.}
\label{fig6}
\end{figure}

\begin{figure*}[htbp]
\centering
\includegraphics[width=1\linewidth]{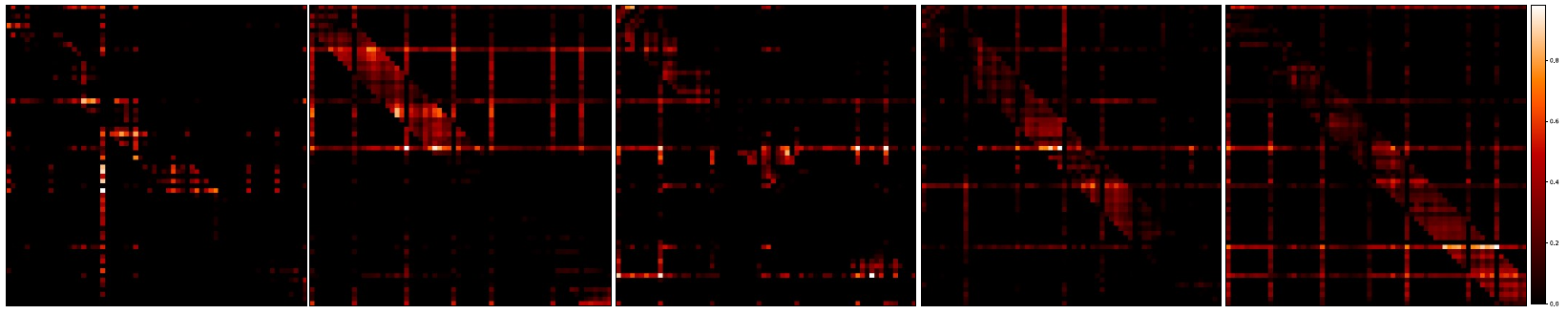}
\caption{Visualization of attention maps of one head for the 26-$th$ video in TVSum at epochs 1, 50, 100, 200, and 300 from left to right, which demonstrates how the attention distribution evolves during training.}
\label{fig7}
\end{figure*}

\begin{figure*}[htbp]
\centering
\includegraphics[width=1\linewidth]{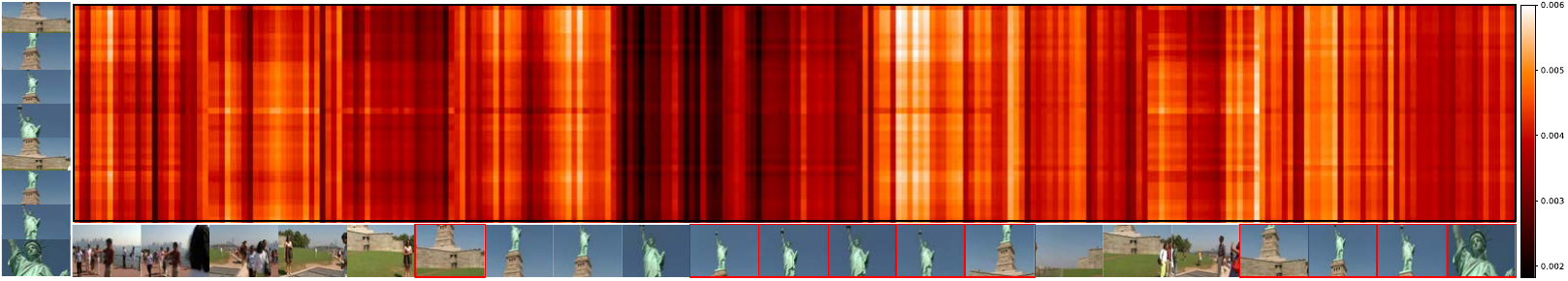}  
\caption{Visualization of the last encoder-decoder attention map generated by FullTransNet for the 20-$th$ video from SumMe.}
\label{fig8}
\end{figure*}

To better understand how attention, particularly local-global sparse attention works, we visualized the attention maps of the encoder, decoder, and encoder-decoder, as shown in Fig. \ref{fig5}, Fig. \ref{fig6}, Fig. \ref{fig7}, and Fig. \ref{fig8}.

In Fig. \ref{fig5}, we visualize the encoder, decoder, and encoder-decoder attention maps for the 12-$th$ video of SumMe, where the attention weights are extracted from the model trained on SumMe dataset for one epoch under the canonical setting. Fig. \ref{fig5}(a), Fig. \ref{fig5}(b), and Fig. \ref{fig5}(c) show the local-global sparse, masked, and cross attention maps, respectively, all with 6 layers and 8 heads. In each subfigure, rows represent layers 0--5 from top to bottom, columns represent heads 0--7 from left to right. 
Taking Fig. \ref{fig5}(a) as an example, it can be seen that different heads in the same row attend to different information, for instance, for layer 4, heads 0, 2, and 7 attend to the information of the same position, while other heads attend to the information of different position; and the information is primarily from the neighborhood of the position. The reason should be attributed to our LGS-MHSA mechanism.

To further demonstrate how one position attends to its neighbors or all corresponding tokens, we visualize the attention maps of LGA, masked, and cross-attention from layer 0 and head 0 for the 8-$th$ video of SumMe. As can be observed from Fig. \ref{fig6}, a grid-like area is highlighted in the LGA attention map, which well fits with the designed attention pattern, that is, the intersecting position in this area indicates that the token at that position attends to all tokens in the sequence, while the highlighting band area shows that each token only attends to the tokens within a fixed window. The qualitative results show that applying local attention to several tokens rather than all tokens is effective. In fact, not all frames need to be attended to. The masked attention map shows that summary frames fed into the decoder present in a lower triangular style, while the cross-attention map shows that the frames being attended to present in a vertical streaks manner. 
This implies that a few positions in the output sequence receive significantly greater weights from all elements in the input sequence.

Since LGA is one of our main technical contributions, to further inspect how the LGA works, we visualize the attention maps of head 0 from layer 0 at epochs 1, 50, 100, 200, and 300 for the 26-$th$ video in TVSum, where the weights are from the models trained on TVSum under the canonical setting. One can see from Fig. \ref{fig7} that the model starts to converge around 200 epochs, with the attention distribution becoming increasingly stable.

We specifically visualize the attention map of the last encoder-decoder layer for the 20-$th$ video from SumMe to demonstrate how the attention map is affected by our LGA, as shown in Fig. \ref{fig8}. The horizontal axis represents the original sequence of the video, i.e., all frames of the video, while the vertical axis represents the target sequence, i.e., the sequence of keyframes. 
One can observe that the keyframes in this exemplar are primarily located in the middle or latter part of the video. These keyframes, serving as the query sequence, effectively focus on video frames that are similar to them (highlighted in red boxes). Some target frames may concentrate on a specific frame in the original video, resulting in a vertical streak through the attention-weight space. This phenomenon may occur because the target frames belong to one or more shots where the content is relatively similar, and the specific frame happens to lie within such a shot. Alternatively, this frame may be highly relevant to the target frame, i.e., a frame within the keyframe sequence.

\section{Conclusion}
\label{sec:conclusion}

In this paper, we propose FullTransNet, a transformer-like network for video summarization. Unlike existing methods, FullTransNet adopts a full transformer architecture with an encoder-decoder structure and incorporates a local-global sparse attention mechanism to achieve comparable or even superior summarization performance. Specifically, the encoder-decoder structure is designed to model both video understanding and summary generation, while the local-global sparse attention mechanism aims to reduce computational costs without sacrificing performance. Experimental results demonstrate the superiority of this full transformer with local-global sparse attention over recently proposed deep models and encoder-only transformer architectures.

While FullTransNet performs well on benchmark datasets such as SumMe and TVSum, it has several limitations. First, FullTransNet is trained on short video clips, which may limit its generalization to longer videos or more complex scenarios. Specifically, videos with rapid scene transitions or multi-shot sequences pose significant challenges for the model. These challenges include maintaining temporal coherence and accurately summarizing diverse content. Second, the current design of FullTransNet may encounter computational bottlenecks when applied to real-time video summarization tasks, particularly for high-resolution or long-duration videos.

To address these limitations, our future work will focus on several key directions. First, we plan to develop more sophisticated sparse attention mechanisms and a generalized architecture for video summarization tasks. These enhancements will enable the model to capture both local details and global context in longer videos, while improving its ability to handle multi-shot scenes and enhance temporal modeling. Second, we aim to incorporate advanced training techniques, such as synthetic data augmentation and pretraining on large-scale video datasets, to improve the model’s robustness and adaptability to varied video content. Finally, to address computational challenges, we will explore lightweight model variants and hardware acceleration techniques to ensure scalability for real-time applications.

\bibliographystyle{IEEEtran}
\bibliography{IEEEabrv,refref}

\end{document}